%% file: main.tex
\documentclass[10pt,twocolumn,letterpaper]{article}

\usepackage{wacv}
\usepackage{times}
\usepackage{epsfig}
\usepackage{graphicx}
\usepackage{amsmath}
\usepackage{amssymb}
\usepackage{tabularx}
\usepackage{url}
\usepackage{color}
\usepackage{xspace}
\usepackage{balance}
\usepackage{enumerate}
\usepackage{multirow}
\usepackage[linesnumbered, lined, ruled]{algorithm2e}
\usepackage[pagebackref=false,breaklinks=true,colorlinks,bookmarks=false,linkcolor=blue,citecolor=blue,pdftitle={},pdfauthor={}]{hyperref}

\DeclareMathOperator*{\argmax}{arg\,max}

\makeatother

\newcommand{\vecz}{\boldsymbol{z}}
\newcommand{\vecalpha}{\boldsymbol{\alpha}}

\newcommand{\image}{\boldsymbol{I}}

\newcommand{\vecy}{\boldsymbol{y}}

\def\wacvPaperID{112} 

\ifwacvfinal\fi
\wacvfinalcopy 

\begin{document}

\title{Fine-Grained Classification via Mixture of Deep Convolutional Neural Networks}

\author{Zongyuan Ge \hspace{2cm} Alex Bewley \hspace{2cm} Christopher McCool\\
Australian Center for Robotic Vision\\
{\tt\small z.ge@qut.edu.au}
\and
Second Author \\
Institution2\\
{\tt\small secondauthor@i2.org}
}

\author{ZongYuan Ge{\tiny ~}$^{\dagger\ddagger*}$, Alex Bewley{\tiny ~}$^{\ddagger}$,Christopher McCool{\tiny ~}$^{\ddagger*}$, Ben Upcroft{\tiny ~}$^{\dagger\ddagger}$,Peter Corke{\tiny ~}$^{\dagger\ddagger}$, Conrad Sanderson{\tiny ~}$^{\ast\diamond}$
\\
$^\dagger$     Australian Centre for Robotic Vision, Brisbane, Australia\\
  $^\ddagger$  Queensland University of Technology (QUT), Brisbane, Australia\\
  $^\ast$ University of Queensland, Brisbane, Australia\\
  $^\diamond$ NICTA, Australia
}


\maketitle
\ifwacvfinal\thispagestyle{empty}\fi

\input{abstract}
\input{intro}

\input{background}

\input{proposed_method}

\input{experiment}

\input{conclusion}

\section*{Acknowledgements}

\begin{small}
The Australian Centre for Robotic Vision is supported by the Australian Research Council via the Centre of Excellence program.
NICTA is funded by the Australian Government through the Department of Communications,
as well as the Australian Research Council through the ICT Centre of Excellence program.
Agricultural Robotics Program at QUT has been supported by the Department of Agriculture 
\end{small}

\balance
{\small
\bibliographystyle{ieee}
\bibliography{refs}
}

\end{document}

%% file: abstract.tex

\begin{abstract}


We present a novel deep convolutional neural network (DCNN) system for fine-grained image classification, called a mixture of DCNNs (MixDCNN).
The fine-grained image classification problem is characterised by large intra-class variations and small inter-class variations.
To overcome these problems our proposed MixDCNN system partitions images into $K$ subsets of similar images and learns an expert DCNN for each subset.
The output from each of the \mbox{$K$~DCNNs} is combined to form a single classification decision.
In contrast to previous techniques, we provide a formulation to perform \textbf{joint} end-to-end training of the \mbox{$K$~DCNNs} simultaneously. 
Extensive experiments, on three datasets using two network structures (AlexNet and GoogLeNet), show that the proposed MixDCNN system consistently outperforms other methods.
It provides a relative improvement of 12.7\% and achieves state-of-the-art results on two datasets.
\end{abstract}

%% file: intro.tex
\section{Introduction}
\label{sec:introduction}

\begin{figure}[!tb]
  \centering
  \includegraphics[width=0.8\columnwidth]{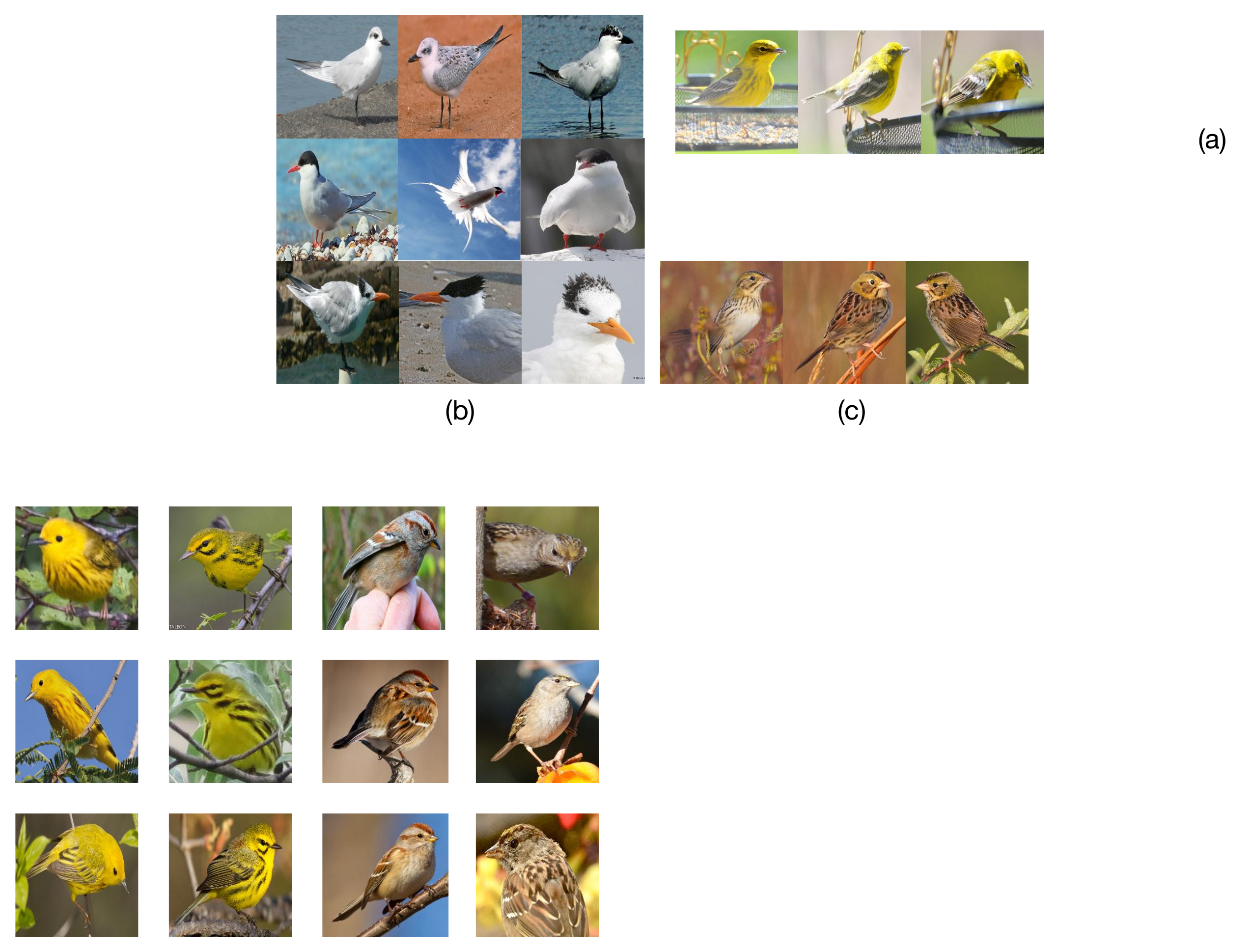}
  \caption{
    Example images from the Birdsnap dataset~\cite{berg2014birdsnap} which exhibits large intra-class variations and low inter-class variations.
    Each column represents a unique class.
    }
  \label{fig:birdsnap}
  \vspace{0.5ex}
  \hrule
  \vspace{-2ex}
\end{figure}

Fine-grained image classification consists of discriminating between classes in a sub-category of objects,
for instance the particular species of bird or dog~\cite{berg2013poof,chai2013symbiotic,farrell2011birdlets,gavves2013fine,zhang2014part}.
This is a very challenging problem due to large intra-class variations (due to pose and appearance changes),
as well as small inter-class variation (due to only subtle differences in the overall appearance between classes).
See Fig.~\ref{fig:birdsnap} for examples.

\footnote{$^*$ Authors contributed equally.}

To cope with the above problems,
many fine-grained classification methods have performed parts detection~\cite{berg2013poof,chai2013symbiotic,liu2012dog,zhang2013deformable}
in order to decrease the intra-class variation.
Recently, an alternative approach was introduced by Ge et al.~\cite{gecvpr2015} where the images were first partitioned into $K$ non-overlapping sets and $K$ expert systems were learned.
By grouping similar images, the input space is being partitioned so that an expert network can better learn the subtle differences between similar samples.
Expert selection was performed by training a dedicated gating network which assigns samples to the most appropriate expert network.
This approach has two downsides.
Firstly, a separate gating network (subset selector) needs to be trained. 
Secondly, the expert networks are trained only to extract features, leaving the final classification to be performed by a linear support vector machine (SVM).


We propose a novel system based on a mixture of deep convolutional neural networks (DCNNs)
that provides state-of-the-art performance along with several important properties.
Similar to Ge et al.~\cite{gecvpr2015}, we partition the data into $K$ non-overlapping sets to learn $K$ expert \mbox{DCNNs}. 
However, unlike~\cite{gecvpr2015}, the classification decision from the each expert is weighted proportional to the confidence of its decision. 
This allows us to define a single network (\mbox{MixDCNN}), comprised of $K$ sub-networks (expert \mbox{DCNNs}), that can be trained to perform classification.
This is in contrast to~\cite{gecvpr2015}, where each expert is used just for feature extraction.
Our system has similarities to the gated network approach proposed by Jacobs et al.~\cite{jacobs1991adaptive},
which utilises a separately trained network to select the most appropriate expert network. 

The proposed MixDCNN system allows us to jointly train the network, which has two advantages:
{\bf (i)} it obviates the need for a separate gating network,
and
{\bf (ii)} samples can be re-assigned to the most appropriate expert network during the training process.
%
Empirical evaluations show that this approach outperforms related approaches such as subset feature learning~\cite{gecvpr2015},
a gated DCNN approach similar to~\cite{jacobs1991adaptive},
and an ensemble of classifiers.

The paper is continued as follows.
In Section~\ref{ref:prior_work} we briefly review recent advances in fine-grained classification
and overview approaches to learn multiple expert classifiers, particularly within the field of neural networks.
In Section~\ref{sec:proposed_method} we present our proposed MixDCNN approach in detail. 
Section~\ref{sec:experiment} is devoted to a comparative evaluation against several recent methods on the task of fine-grained classification.
Conclusions and possible future avenues of research are given in Section~\ref{sec:conclusion}.

%% file: background.tex
\section{Prior Work}
\label{ref:prior_work}



Prior work for fine-grained image classification has concentrated on performing parts detection~\cite{berg2013poof,chai2013symbiotic,liu2012dog,zhang2013deformable} in order to decrease the intra-class variation.
The part-based one-vs-one feature system~\cite{berg2013poof} is an example of this, where parts-based features are progressively selected to improve classification.
An alternative is the deformable parts-model which obtains a combined feature from a set of pre-defined parts~\cite{zhang2013deformable}.
Chai et al.~\cite{Chai13_1:conference} proposed a symbiotic model where part localisation is helped by segmentation and, conversely, the segmentation is helped by parts detection. 
Zhang et al.~\cite{zhang2013deformable} extract pose-normalised features based on weak semantic annotations to learn cross-component correspondences of various parts.

Recent work has shown the effectiveness of DCNNs for fine-grained image classification, but again, predominantly to perform parts detection.
Region proposal methods combined with a DCNN were shown to more accurately localise object parts~\cite{zhang2014part}. 
Lin et al.~\cite{lin2015deep} showed that a DCNN can be trained to perform both parts localisation and visibility prediction,
achieving state-of-the-art results on the CUB dataset~\cite{wah2011caltech}.  
Although the above parts-based approaches are fully automatic at test time, they require a large number of images to be manually annotated in order to train the model.

To remove the need for time-consuming manual annotations, recent work has explored ways to perform fine-grained classification without using part annotations.
Zhang et al.~\cite{zhang2014part} and Ge et al.~\cite{ge2015modelling} showed that, even without part annotations,
\mbox{DCNNs} can provide impressive performance for fine-grained classification tasks.
Of particular interest is the approach of Ge et al.~\cite{Ge_ICIP_15} which showed that the data can be partitioned into $K$ non-overlapping sets
and an expert feature extraction algorithm, utilising DCNNs, can be trained for each of the $K$ sets.

Learning algorithms which construct a set of $K$ classifiers and make decisions by taking a weighted or average of their predictions are often referred to as ensemble methods. 
A simple ensemble approach called {\it bagging} has been used to improve the overall performance of a system~\cite{breiman1996bagging}.
Bagging manipulates the training examples to generate multiple hypotheses.
In this case, a set of $K$ classifiers is learned using a randomly selected subset of the training data.
We use this bagging approach on a set of DCNNs for a baseline method and refer to it as an Ensemble approach (Section \ref{sec:experiment}).

Ensemble approaches, or learning $K$ expert classifiers, has been explored by several researchers within the context of neural networks.
In 1991 Jacobs et al.~\cite{jacobs1991adaptive} described a gated network structure to learn $K$ expert neural networks and applied it to multi-speaker vowel recognition.
The underlying idea is to only allocate a small region of the input space to a particular expert system. 
This was achieved by having $K$ expert systems (neural networks) which were allocated samples selected by a separate gating network.
In~\cite{jacobs1991adaptive}, the gating network determines the probability that a sample is associated to one of the $K$ expert systems.


More recently, Ge et al.~\cite{gecvpr2015} outlined a subset feature learning (Subset FL) approach using $K$ expert DCNNs.
The data is partitioned into $K$ non-overlapping sets and for each set an expert DCNN is learned to extract set-specific features.
A gating network is then used to extract only the most relevant features from these $K$ DCNNs.
Classification is then performed by training an SVM on these features,
yielding impressive performance for fine-grained bird and plant classification~\cite{Ge15_1:competition}.
An issue with this work is the reliance of an independent gating network $\mathcal{G}$
and the fact that feature extraction and classification are treated as independent steps.

%% file: proposed_method.tex
\section{Proposed Approach}
\label{sec:proposed_method}

\begin{figure*}[!tb]
  \centering
  \includegraphics[width=0.8\textwidth]{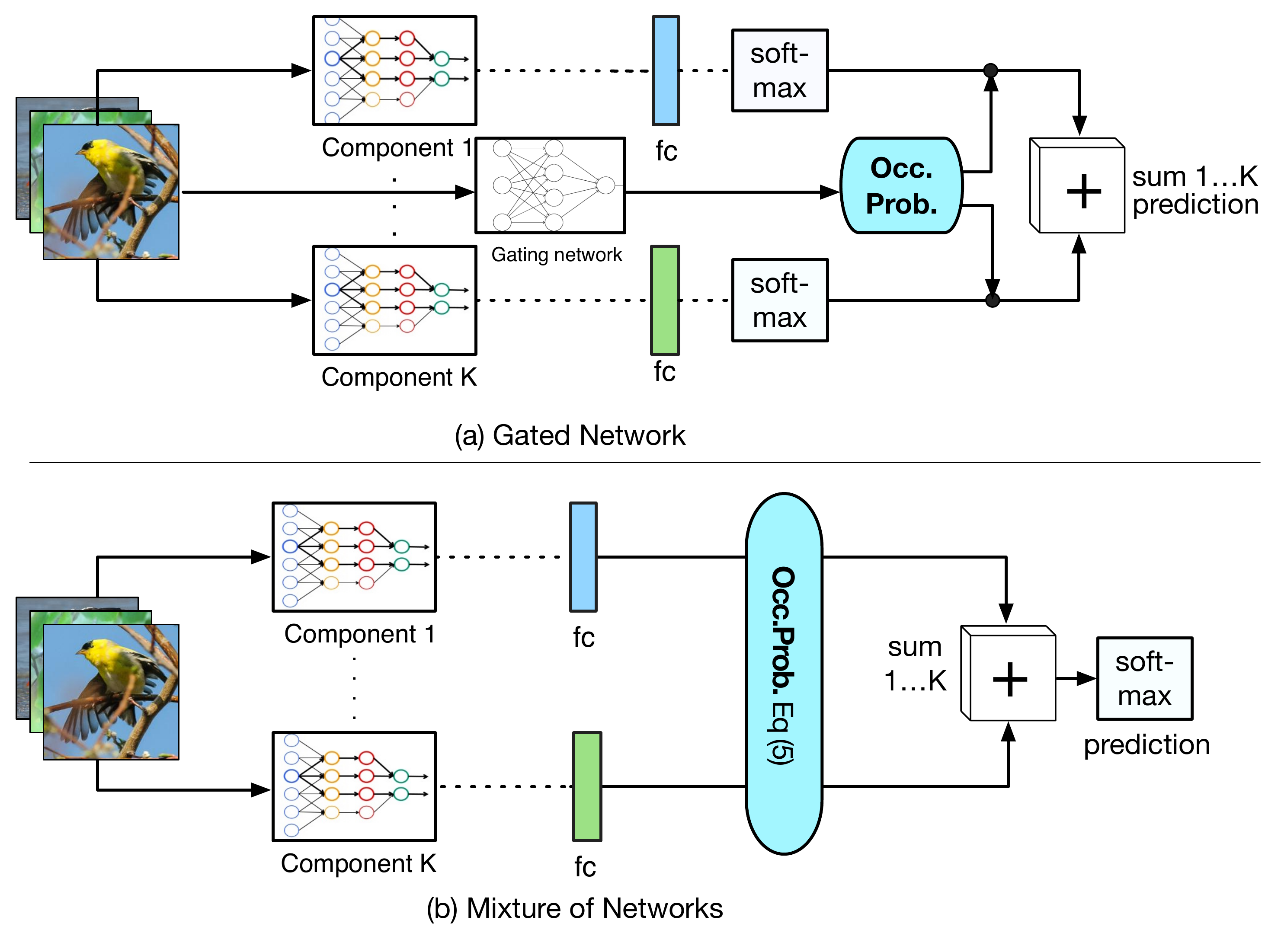}
  \caption{
    GatedDCNN structure (top) and MixDCNN structure (bottom). 
    The term {\it Occ.~Prob.} refers to occupation probability (responsibility) $\vecalpha$. 
    In GatedDCNN, the gating network uses the image, the same input as each component (subset networks), to estimate $\vecalpha$.
    In~contrast, MixDCNN estimates $\vecalpha$ without the need for an external network.
    }
  \label{fig:network}
\end{figure*}

We propose a novel mixture of DCNNs (MixDCNN) to improve fine-grained image classification by partitioning the data into $K$ non-overlapping sets and learning an expert classifier for each set.
This approach has similarities to the gated neural network proposed by Jacobs et al.~\cite{jacobs1991adaptive}, which has never been applied to DCNNs nor to the fine-grained classification problem.
As such, we also outline a gated DCNN (GatedDCNN).
An overview of these two approaches is given in Figure~\ref{fig:network}.

The main idea behind the MixDCNN and GatedDCNN approaches is to learn $K$ expert networks, $\left[ \mathcal{S}_{1}, \dots, \mathcal{S}_{K} \right]$, which make decisions about a subset of the data.
This simplifies the space that is being modelled by each component.
Key to both approaches is being able to assign a sample to the appropriate network.

A GatedDCNN assigns samples by learning a separate gating neural network which produces the probability, $\alpha_{k}$, that the sample belongs to the $k$-th network.
Learning this gating neural network requires ground truth labels about which sample should be assigned to a particular network, which for our work is an open question.
In contrast, a \mbox{MixDCNN} assigns samples based on the confidence of the prediction from each network,
which leads us to consider $\alpha_{k}$ to be the occupation probability of the sample for the $k$-th network.

Before we describe these two approaches in more detail we define some notation.
The output of a DCNN, trained for classification, is an $N$-dimensional vector $\vecz$ of class predictions, where $N$ indicates the number of classes.
These predictions then are normalised by a softmax~\cite{krizhevsky2012imagenet,szegedy2014going} to give the probability that the sample belongs to the $n$-th class:
\begin{equation}\label{eq:softmax}
  c_n = \frac{\exp\{z_{n}\}}{\sum_{j=1}^{N}\exp\{z_{j}\}}
\end{equation}
In the approaches described below, we are most interested in the vector of predictions $\vecz$ prior to applying the softmax.

\subsection{GatedDCNN}
\label{sec:gated}

Inspired by~\cite{hampshire1992meta,jacobs1991adaptive}, we define a GatedDCNN that consists of $K$ components (DCNNs) and an additional gating network.
The overall structure of this network is shown in Fig.~\ref{fig:network}a.
In this arrangement, the $k$-th DCNN $\mathcal{S}_{k}$ is given greater responsibility for learning to discriminate subtle differences of the $k$-th subset of images,
while the gating network $\mathcal{G}$ is responsible for associating the image $\image$ with the most appropriate component.
The gating network $\mathcal{G}$ is a fine-tuned DCNN that is learned using the cross-entropy loss to produce a $K$-dimensional vector of probabilities $\vecalpha$.
The $k$-th value denotes the probability that the input image $\image$ is associated with the $k$-th component.
We refer to this as an occupation probability.


A fundamental difficulty with training the GatedDCNN is how to provide the $T$ training labels $\vecy$. This label vector is a $K$-dimensional label vector which indicates which of the $K$ subsets the sample belongs to. 
To deal with this issue we consider two ways of estimating these labels.
The first approach is to initialise the labels $\vecy$ using the partitioning of the training images into $K$ subsets.
The gated network $\mathcal{G}$ is then trained using these labels and the $K$~\mbox{DCNNs} (components) are then trained independently so that $\mathcal{S}_{k}$ is trained exclusively with data from the $k$-th subset.
The second approach is to use the above gated network (and $K$ components) as an initialisation and to iteratively retrain by: 
\begin{enumerate}
  \item Fixing $\mathcal{G}$, and then updating $\left[ \mathcal{S}_{1}, \dots, \mathcal{S}_{K} \right]$ using the assignments from $\mathcal{G}$.
  \item Fixing the $K$ components $\left[ \mathcal{S}_{1}, \dots, \mathcal{S}_{K} \right]$ and using these to estimate new labels $\vecy$. The network $\mathcal{G}$ is then updated using these new labels. 
\end{enumerate}

The labels $\vecy$ estimated in step 2 are obtained by taking the network which is most confident about its decision.
Formally, $y_{t}$ for the $t$-th training sample is given by:
\begin{equation}
  y_{t} = \underset{k = 1 \dots K} \argmax \; C_{k,t}
\end{equation}
where $C_{k,t}$ is the best classification result for $\mathcal{S}_{k}$ using the $t$-th sample:
\begin{equation} \label{eq:best_classification}
  C_{k,t} = \underset{n= 1 \dots N} \max z_{k,n,t}
\end{equation}

Classification with the GatedDCNN is performed using a weighted summation of the classification results from the $K$ components:
\vspace{-2ex}
\begin{equation}\label{eq:gated_weighting}
	c_{n} = \sum\nolimits_{k=1}^{K} c_{k,n}\alpha_{k}
\end{equation}
where $c_{k,n}$ is the probability of the sample belonging to the $n$-th class for the $k$-th component,
and $\alpha_{k}$ is the probability that the sample is assigned to the $k$-th component $\mathcal{S}_k$.

An issue with the GatedDCNN system is that a separate gating network has to be trained to assign a sample to a particular component $\mathcal{S}_k$.
This provides the further complication of having to estimate the labels $\vecy$ in order to train the gating network $\mathcal{G}$.
In this paper the first GatedDCNN training approaches provides marginally better performance. In the experiment section, we will report results based on the first approach.

  
 



\subsection{Mixture of DCNNs (MixDCNN)}
\label{sec:mixture}

We propose a mixture of DCNNs approach where the occupation probabilities $\vecalpha$ are based on the classification confidence from each component.
An advantage of this structure is that we can jointly train the $K$~DCNNs (components) without having to estimate a separate label vector $\vecy$ or train a separate gating network $\mathcal{G}$.


For MixDCNN, the occupation probability for the \mbox{$k$-th} component is:
\vspace{-2ex}
\begin{equation}\label{eq:occ_prob}
  \alpha_{k} = \frac{\exp\{C_{k}\}}{\sum_{c=1}^{K} \exp\{C_{c}\}}
\end{equation}
where $C_{k}$ is given by Eq.~\eqref{eq:best_classification}.
This occupation probability gives higher weight to components that are confident about their prediction.
The overall structure of this network is shown in Fig.~\ref{fig:network}b.

Classification is performed by multiplying the output of the final layer from each component by the occupation probability and then summing over the $K$ components:
\begin{equation}\label{eq:mix_weighting}
  z_{n} = \sum\nolimits_{k=1}^{K}z_{k,n}\alpha_{k}
\end{equation}
This mixes the network outputs together and the probability for each class is then produced by applying the softmax function in Eq.~\eqref{eq:softmax}.
As a consequence our MixDCNN is optimised using the cross-entropy loss\footnote{Optimised in a mini-batch Stochastic Gradient Descent framework.}.



\subsection{Differences Between MixDCNN and Ensembles}

The aim of the MixDCNN approach is that each component takes greater responsibility for a portion of the data allowing each component to concentrate on samples (or classes) that are more difficult to differentiate.
This will allow the MixDCNN to learn subtle differences for similar classes. 
This is in contrast to an Ensemble approach which randomly excludes a portion of the training data for each DCNN.
Therefore, the key difference between the proposed MixDCNN approach and an ensemble of DCNNs (Ensemble) is the use of the occupation probability.
For training, this means the MixDCNN approach does not randomly select the data.
Instead, each sample is weighted proportional to its relevance to each DCNN $\mathcal{S}_{1, \dots, K}$.
For testing, the MixDCNN approach is able to adaptively calculate the occupation probability for each sample,
whereas an Ensemble approach will use pre-defined weights or, more commonly, equal weights.

%% file: experiment.tex
\section{Experiments}
\label{sec:experiment}

\subsection{Datasets}

\begin{figure}[!b]
  \centering
  \includegraphics[width=1\columnwidth]{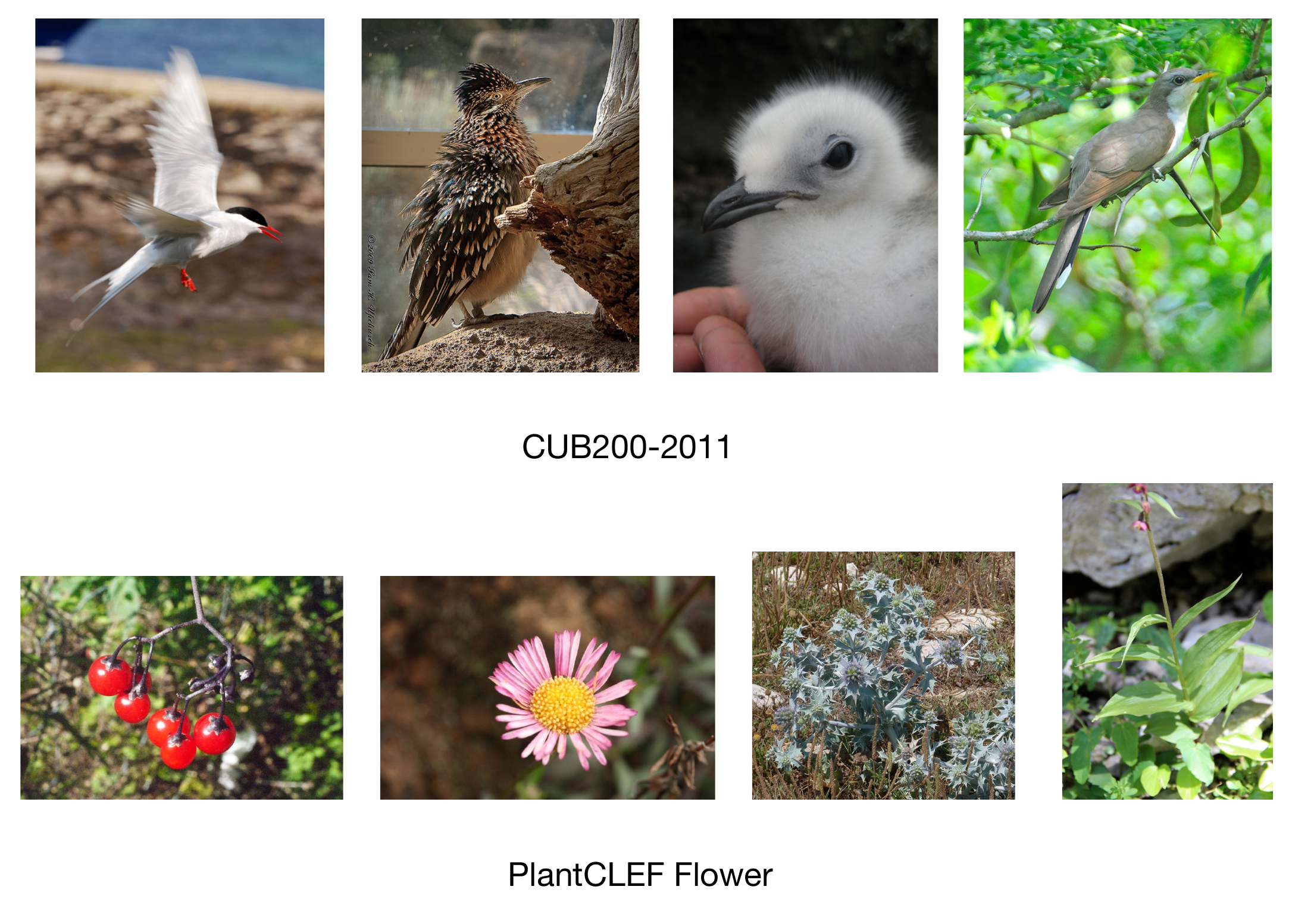}
  \caption
    {Examples from CUB-200-2011 and PlantCLEF Flower. 
    }
  \label{fig:cub_flower}
\end{figure}

We present results on three fine-grained image classification datasets using two network structures.
The three datasets are the Caltech-UCSD-2011 (CUB200-2011)~\cite{wah2011caltech}, Birdsnap~\cite{berg2014birdsnap}, and PlantCLEF 2015~\cite{goeau2014lifeclef}.
Example images are shown in Figures~\ref{fig:birdsnap} and~\ref{fig:cub_flower}.

CUB200-2011 is a fine-grained bird classification task with 11,788 images from 200 bird species in North America.
This dataset has become a \textit{de facto} standard for the bird classification task.
Each species has approximately 30 images for training and 30 for testing.
Birdsnap is a much larger bird dataset consisting of 49,829 images from 500 bird species with 47,386 images used for training and 2,443 images used for testing.
PlantCLEF 2015 is a large plant classification dataset that has seven content types.
To demonstrate the capabilities of the proposed \mbox{MixDCNN} approach for the task of fine-grained classification,
we analyse its effectiveness on one content type, Flower.
This portion of the dataset consists of 28,705 images from 967 species.
We split this data into training and test sets.
The training set consists of 25,025 images from 967 species, while the test set has 3,200 images from 801 species.

Both \mbox{CUB200-2011} and Birdsnap have bounding box annotations around the object of interest.
We use this information to extract just the object of interest from the image.
PlantCLEF 2015 does not come with bounding box information making it a more challenging dataset.

Prior work~\cite{gecvpr2015,zhang2014part} has shown the importance of transfer learning for the fine-grained image classification problem. 
Results have shown that training a DCNN from scratch for either the fine-grained CUB200-2011 or Birdsnap dataset leads to overfitting on the training samples. 
As such, for all the of our experiments we use pre-trained networks from ImageNet~\cite{deng2009imagenet}
to provide a good initialisation for each DCNN and then perform transfer learning. 
We consider this to be our baseline and refer to it as \textbf{DCNN-tl}.
All of our networks are trained using Caffe~\cite{jia2014caffe} and partitioning was performed using the Bob toolkit~\cite{bob2012}.


\subsection{Comparative Evaluation}

\begin{table*}[!tb]
\centering
\caption
  {Comparison of the proposed MixDCNN approach against DCNN-tl, Ensemble, GatedDCNN and Subset FL on three datasets: CUB, BirdSnap and PlantCLEF-Flower.
  Two network structures are used: AlexNet and GoogLeNet.
 \vspace{1ex}  
  }
\scalebox{1.0}
  {
  \begin{tabular}{l|c|c|c|c|c|c}
  \multicolumn{2}{c|}{}               &  \textbf{DCNN-tl}   & \textbf{Ensemble}    & \textbf{GatedDCNN}   & \textbf{Subset FL}    & \textbf{MixDCNN} \\ \hline
  \multirow{3}{*}{\textbf{AlexNet}}   & CUB        & 68.3$\%$    & 71.2$\%$     & 69.2$\%$             & 72.0$\%$                  & \textbf{73.4$\%$}         \\ \cline{2-7} 
                                    & BirdSnap   & 55.7$\%$    & 57.2$\%$                 & 57.4$\%$             & 59.3$\%$                  & \textbf{63.2$\%$}          \\ \cline{2-7}
                                  & PlantCLEF-Flower     & 29.1$\%$    & 30.2$\%$                 & 30.2$\%$             & 31.1$\%$                  & \textbf{35.0$\%$}         \\ \hline \hline 

  \multirow{3}{*}{\textbf{GoogLeNet}} & CUB        & 80.0$\%$    & 80.9$\%$                 & 81.0$\%$             & \textbf{ 81.2$\%$}                  & 81.1$\%$         \\ \cline{2-7} 
                                      & BirdSnap   & 67.4$\%$    & 71.4$\%$                & 70.1$\%$             & 72.8$\%$                  & \textbf{74.1$\%$}         \\ \cline{2-7}
                                      & PlantCLEF-Flower     & 48.7$\%$    & 50.2$\%$                 &49.7$\%$              & 51.7$\%$                  & \textbf{52.1$\%$}         \\ \hline 
  \end{tabular}
  }
 \label{table:re_2}
 \vspace{1ex}
\end{table*}

We compare the proposed MixDCNN approach against four other related methods:
{\bf (1)}~the baseline DCNN-tl,
{\bf (2)}~an ensemble of $K$ DCNNs,
{\bf (3)}~an implementation of GatedDCNN,
and
{\bf (4)}~Subset~FL~\cite{gecvpr2015}.
Two network structures considered are the well known AlexNet~\cite{krizhevsky2012imagenet}
and the Large Scale Visual Recognition Challenge (ILSVRC) 2014 winner GoogLeNet~\cite{szegedy2014going}.
AlexNet is a deep network consisting of 8 layers,
while ILSVRC has 22 layers\footnote{To prevent GoogLeNet from over-fitting we use a higher dropout rate equal to 0.5 for the final loss layer, as opposed to the original setting of 0.4.}.
We follow the same procedure as Ge et. al~\cite{gecvpr2015} to cluster the data.
For the AlexNet structure we use the output of the first fully connected layer as features for clustering.
For GoogLeNet we use the output of the last layer, prior to classification, as features.
In both cases linear discriminant analysis (LDA) is applied to reduce the dimensionality to $D=128$. In our initial experiments, we varied $D$ and results showed no impact of that.
 
The results in Table~\ref{table:re_2} show that the proposed \mbox{MixDCNN} approach provides consistent improvement regardless of network structure or dataset.
MixDCNN provides the best performance for all of the network and dataset combinations, with the exception of the MixDCNN model using the GoogLeNet structure on CUB.
It provides an average relative performance improvement of 12.7\% over the baseline DCNN-tl approach, excluding CUB.

For the CUB dataset, using multiple expert networks provides limited performance improvement.
This is true for all of the methods examined.
We attribute this to the fact that CUB200-2011 is a small dataset consisting of just 5,994 training images.
This is an order of magnitude fewer samples than other datasets such as Birdsnap.
Furthermore, applying transfer learning to GoogLeNet already provides exceptional performance and so minimises the improvement introduced by the MixDCNN framework, or any multi-expert approach.

The proposed \mbox{MixDCNN} method achieves state-of-the-art results on the challenging Birdsnap and PlantCLEF-Flower datasets.
For Birdsnap the previous state-of-the-art performance was 48.8\%~\cite{berg2014birdsnap}. 
Applying transfer learning to GoogLeNet already outperforms this prior art with an accuracy of 67.4\%.
MixDCNN provides a further relative performance improvement of 9.9\%.
For the PlantCLEF-Flower dataset the baseline performance of DCNN-tl (using GoogLeNet) is 48.7\%.
MixDCNN provides state-of-the-art performance with a relative performance improvement of 7.0\%.

The MixDCNN approach consistently outperforms the Ensemble, GatedDCNN and Subset FL approaches. 
Interestingly, it provides a considerable improvement over the closely related GatedDCNN approach, with an average relative performance improvement of 9.1\%.
We attribute this to the ability of the MixDCNN approach to adaptively re-assign samples to the most appropriate expert network, in spite of the original partitioning.

In our experiments, component sizes greater than $K=6$ were not considered as we could not store these in memory on a single GPU\footnote{The GPU used in all our experiments was an Nvidia K40 Tesla with 12~Gb of memory.}.
This highlights one of the limitations with this technique as it currently requires all of the networks to be stored on a single GPU; future work should consider how to extend the architecture across multiple GPUs.

%% file: conclusion.tex
\section{Conclusion}
\label{sec:conclusion}

We have proposed a novel mixture of deep neural networks, termed MixDCNN, which achieves state-of-the-art performance for fine-grained classification.
It provides an average relative performance improvement of 12.7\%
and has been shown to consistently outperform several related methods: subset feature learning, GatedDCNN, and an ensemble of classifiers.

The key advantage of our proposed approach is the use of an occupation probability that weights each sample proportional to its relevance to each DCNN $\mathcal{S}_{1, \dots, K}$.
This approach obviates the need for a separate gating function and highlights the importance of being able to adaptively weight samples based on their relevance to a component (DCNN).

%
%

%

Future work will explore alternative methods for initialising the clustering and its impact upon performance. 
For instance, the impact of grouping images together in terms of their pose rather than similar visual appearance.
Furthermore, we will examine the role of the occupation probability in two ways:
{\bf (i)}~whether the responsibility for a sample is shared between components,
and 
{\bf (ii)}~deeper analysis of how this occupation probability changes during the training process.
Additionally, we intend on exploring different methods for computing the occupational probability via alternative aggregation techniques.

%
%
%
%
%
%

%